\documentclass{ecai} 
\usepackage{latexsym}
\usepackage{amssymb}
\usepackage{amsmath}
\usepackage{amsthm}
\usepackage{booktabs}
\usepackage{enumitem}
\usepackage{graphicx}
\usepackage{color}
\usepackage{apalike}

\newcommand{\BibTeX}{B\kern-.05em{\sc i\kern-.025em b}\kern-.08em\TeX}

\begin{document}

%%%%%%%%%%%%%%%%%%%%%%%%%%%%%%%%%%%%%%%%%%%%%%%%%%%%%%%%%%%%%%%%%%%%%%%%

\begin{frontmatter}

%%% Use this command to specify your submission number.
%%% In doubleblind mode, it will be printed on the first page.

\paperid{123} 

%%% Use this command to specify the title of your paper.

\title{Model Science: getting serious about verification, explanation and control of AI systems}

%%% Use this combinations of commands to specify all authors of your 
%%% paper. Use \fnms{} and \snm{} to indicate everyone's first names 
%%% and surname. This will help the publisher with indexing the 
%%% proceedings. Please use a reasonable approximation in case your 
%%% name does not neatly split into "first names" and "surname".
%%% Specifying your ORCID digital identifier is optional. 
%%% Use the \thanks{} command to indicate one or more corresponding 
%%% authors and their email address(es). If so desired, you can specify
%%% author contributions using the \footnote{} command.

\author[A]{\fnms{Przemyslaw}~\snm{Biecek}\orcid{0000-0001-8423-1823}\thanks{Corresponding Author. Email: przemyslaw.biecek@pw.edu.pl}}
\author[B]{\fnms{Wojciech}~\snm{Samek}\orcid{0000-0002-6283-3265}}

\address[A]{Centre for Credible AI, Warsaw University of Technology, University of Warsaw}
\address[B]{Department of Artificial Intelligence,
Fraunhofer Heinrich Hertz Institute, Technical University of Berlin, Germany BIFOLD - Berlin Institute for the
Foundations of Learning and Data}

%%% Use this environment to include an abstract of your paper.

\begin{abstract}
The growing adoption of foundation models calls for a paradigm shift from Data Science to Model Science. Unlike data-centric approaches, Model Science places the trained model at the core of analysis, aiming to interact, verify, explain, and control its behavior across diverse operational contexts. This paper introduces a conceptual framework for a new discipline called Model Science,  along with the proposal for its four key pillars: Verification, which requires strict, context-aware evaluation protocols; Explanation, which is understood as various approaches to explore of internal model operations; Control, which integrates alignment techniques to steer model behavior; and Interface, which develops interactive and visual explanation tools to improve human calibration and decision-making. The proposed framework aims to guide the development of credible, safe, and human-aligned AI systems.

\end{abstract}

\end{frontmatter}

%%%%%%%%%%%%%%%%%%%%%%%%%%%%%%%%%%%%%%%%%%%%%%%%%%%%%%%%%%%%%%%%%%%%%%%%

\section{From Data Science to Model Science}

The emergence of \textit{Data Science} as a distinct discipline is deeply rooted in the intellectual developments of the second half of the 20\textsuperscript{th} century. Two landmark works shaped the foundation of this transformation are: John W. Tukey's 1962 essay \textit{``The Future of Data Analysis''} \cite{tukey1962future} and William S. Cleveland's 2001 paper \textit{``Data Science: An Action Plan for Expanding the Technical Areas of the Field of Statistics''} \cite{cleveland2001data}. Together, these works redefined the scope of statistical practice and laid the groundwork for what is now considered modern data science.

Tukey argue that {data analysis should be treated as an autonomous scientific discipline}, distinct from mathematical statistics. In his view, analysis was not merely the application of statistical tests, but a broader process encompassing data collection, exploratory examination, and the iterative refinement of research questions. He advocated for {exploratory data analysis} (EDA) as a central paradigm, stressing that it is often preferable to obtain ``an approximate answer to the right question rather than an exact answer to the wrong one.'' Tukey foresaw the transformative role of {computers and graphical methods}, predicting that advances in computation and visualization would reshape how scientists extract knowledge from data. His ideas encouraged statisticians to engage directly with real-world problems, collaborating with domain experts and embracing the growing complexity and volume of data.

\begin{figure}[t!]
    \centering
     \includegraphics[width=0.38\textwidth]{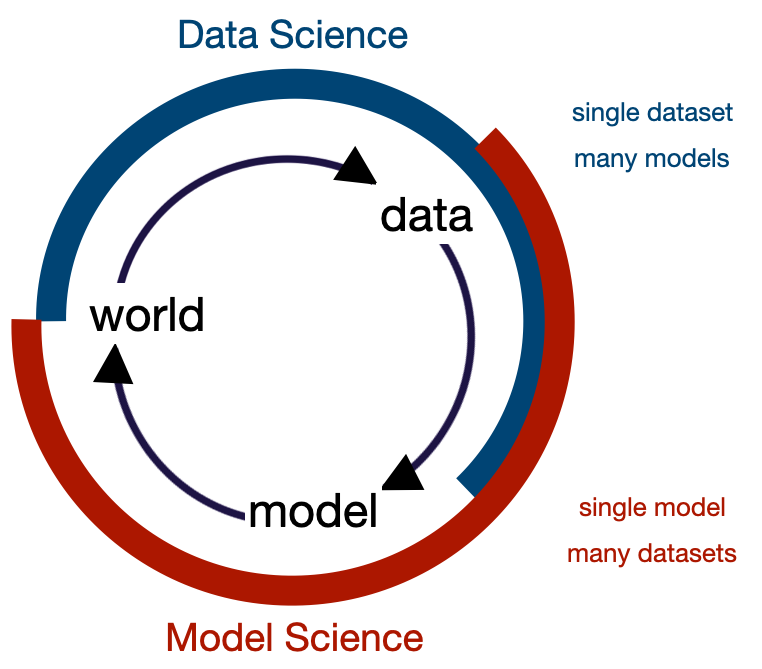} 
      \caption{Data Science and Model Science cover similar modeling areas but emphasize complementary perspectives. Adopting a triadic view, where to understand a phenomenon (\textit{World}) we collect \textit{Data}, build a \textit{Model} based on it, and confront the model with \textit{World} again, Data Science focuses primarily on data. Around a given resource—typically a single well-defined dataset—many models are created to fit the data. Central to this perspective are the data themselves and challenges such as storage, acquisition, visualization, and exploration. In contrast, the Model Science perspective places the chosen model at the center of attention. The analysis of this model may involve multiple datasets (training, validation, monitoring, or out-of-domain), depending on the needs, but all are used primarily to study and understand the model.}
    \label{fig:model_science_intro}
\end{figure}

Nearly four decades later, Cleveland extended Tukey's vision by providing a concrete institutional and methodological framework for the new discipline. He proposed renaming the expanded field of statistics as {``Data Science''}, arguing that traditional statistics should evolve to meet the demands of large, complex, and interdisciplinary data problems. Cleveland proposed a curriculum for Data Science based on six-areas: (1) multidisciplinary investigations, (2) models and methods for data, (3) computing with data, (4) pedagogy, (5) tool evaluation, and (6) theory. This plan explicitly integrated {computing, software development, and algorithmic thinking} as core components of data science and prioritized collaboration with domain scientists as a fundamental driver of methodological innovation. Cleveland emphasized that every technical contribution should ultimately be evaluated by its direct benefit to the practicing data analyst.

The development of Data Science has been driven by the growing capabilities to collect and process large datasets (Big Data). The discipline adopted a highly pragmatic approach to the technical aspects of data cleaning, efficient storage, and data management. Its primary driving force has been data—cheap to collect, process, analyze, and extract value from.

We are now on the brink of another revolution, one driven by large models (Big Models). Large not only in terms of the number of parameters (rapidly exceeding tens of gigabytes) and the number of users (reaching millions for a single model), but also in terms of the range of tasks that can be built upon a single base model (so-called foundation models or frontier models). This paper aims not only to discuss whether a new scientific discipline or sub-discipline—Model Science—might emerge around these models, but also to reflect on the major challenges this discipline will need to address.

Figure \ref{fig:model_science_intro} illustrates the complementary nature of Data Science and Model Science in building models that describe and replicate specific phenomena and processes in the real world, based on collected data, trained algorithms, and the confrontation of these algorithms with reality. The central focus of Data Science is the data itself, along with the challenges of collecting, cleaning, and selecting it. \textit{Data is the King} is the guiding principle of this discipline, where problems are typically formulated around a single well-defined dataset (or a collection of datasets) describing a given phenomenon, on which multiple models are built—usually to select the best-performing one.

The nature of Model Science is complementary, as it focuses on a specific model that is subjected to analysis. The model is the central object, while data becomes a variable: training, validation, test, monitoring, or out-of-domain datasets may not even be fully defined at the training stage but become crucial for verification or analysis. In Model Science, data does not necessarily have to come from real-world measurements; it can also consist of synthetically generated samples, used for additional validation, explanation, or model improvement. The constant element is {the model} under investigation.

Of course, comprehensive analysis often requires a solid understanding of both the data and the model; therefore, Data Science and Model Science frequently overlap—although they pursue different goals and employ different tools. The following sections discuss four main pillars of Model Science.

\section{Four Pillars of Model Science}

Since Model Science focuses on an already trained model, its primary interests concern what can be done after the model has been developed: verification, exploration, control, and communication.

In model analysis, the constant reference point is the model itself, or sometimes several trained models that are compared against each other. The variables in this analysis are the data—these may include data available during training, data generated later (e.g., in out-of-time or out-of-place scenarios), or even synthetic data.

\begin{figure}
    \centering
    \includegraphics[width=0.99\linewidth]{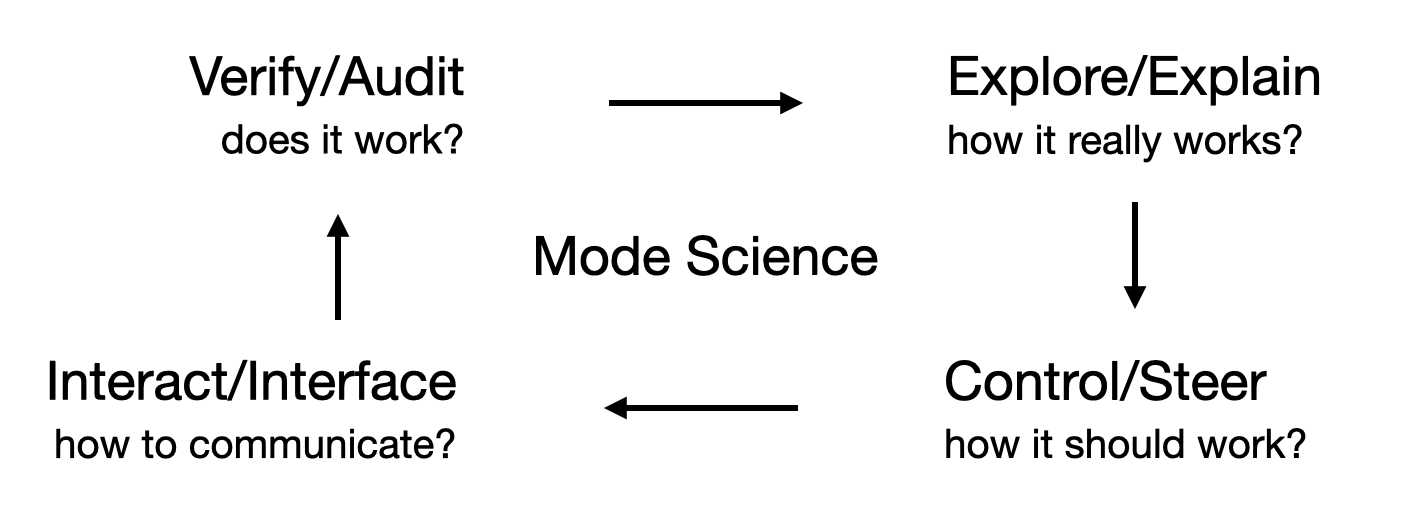}
    \caption{Four main pillars of model analysis}
    \label{fig:model_science_pillars}
\end{figure}

Figure~\ref{fig:model_science_pillars} graphically summarizes the four main pillars of model analysis, which will be discussed in greater detail in the following sections.

\section{Verification: does it work?}

We will begin this section by discussing recently discovered weaknesses in the most popular AI models currently available. Certainly, newer versions of these models will fix (some of) these errors over time, and new errors will appear in newer versions of these models. For this reason, the discussion below does not claim to be universal, but only to emphasize the role of verification, which is neither obvious nor taken seriously enough.

\textbf{Is the AI model evaluation broken?}
AI models and in particular foundation models (FMs) such as large language models (LLMs), speech-to-text (STT) systems, and text-to-image generators are increasingly deployed in domains where errors may have significant ethical, legal, and social consequences. Although these models achieve state-of-the-art (SOTA) performance on standard benchmarks, recent studies reveal that their real-world behavior can deviate significantly from reported metrics, with additional validation often exposing severe flaws. We start with a short review of recent evidence of underperformance and harmful biases in high-stakes contexts, highlighting the importance of rigorous verification protocols prior to deployment. Please note that the cases mentioned below are not intended to stigmatize a particular research group or company for providing an erroneous model; this happens everywhere regardless of the amount of resources involved. This revision is intended to emphasize that systematic research is ongoing to verify large and popular models.

The study \textit{``Careless Whisper: Speech-to-Text Hallucination Harms''}~\cite{koenecke_careless_2024} demonstrates that OpenAI's Whisper---a widely used STT model---produces \emph{hallucinated} sentences absent from the original audio in approximately 1\% of transcripts. Alarmingly, 38\% of these hallucinations contained harmful or violent language. The issue was exacerbated for people with aphasia, where the model incorrectly ``filled in'' long pauses with fabricated statements. These findings reveal the need for domain-specific auditing, as such hallucinations could lead to false legal evidence or medical miscommunication.

Similarly, LLMs have been shown to produce high rates of \emph{legal hallucinations}. The paper \textit{``Large Legal Fictions''}~\cite{dahl_large_2024} reported that ChatGPT-3.5, PaLM-2, and LLaMA-2 invented non-existent case law in 69--88\% of tested legal queries. Even specialized tools such as Lexis+ AI hallucinated in 17--34\% of responses. This contradicts news-media articles of LLMs passing bar exams, underscoring the mismatch between benchmark performance and real-world reliability.

No area is free from these types of problems. In healthcare, a study focused on evaluation of the GPT-4 published in Lancet Digital Health \cite{zack_assessing_2024} revealed persistent {gender and racial biases} in this model. In 37\% of paired clinical scenarios differing only by gender or race, GPT-4 produced distinct diagnoses or treatment suggestions. The model overestimated sarcoidosis prevalence in Black patients and hepatitis B in Asian patients reflecting past historical biases in healthcare. Such errors pose serious risks when models are integrated into diagnostic workflows.

A large-scale analysis of GitHub projects~\cite{majdinasab_assessing_2024} demonstrated that over 35\% of code suggestions generated by GitHub Copilot contained security weaknesses, from diverse group of 42 distinct MITRE’s Common Weakness Enumerations (CWEs), including command injection and insecure randomness. Eleven of these ranked in the CWE Top-25 most dangerous vulnerabilities. This raises concerns about automated code generation in safety-critical software (e.g., medical devices or financial systems).

In education, ChatGPT-3.5 failed to generate reliable medical multiple-choice questions, with only 32\% of items being fully correct and explanatory~\cite{ngo_chatgpt_2024}. Using such unverified material risks propagating misconceptions among medical students. Moreover, audits of LLMs and text-to-image models revealed \emph{adultification bias}~\cite{castleman_adultification_2025}, where Black girls were depicted as older and more sexualized than White peers, potentially reinforcing harmful stereotypes in educational or disciplinary systems.

\textbf{Adversarial analysis} of models and their explanations has become increasingly recognized as essential for ensuring reliability in AI systems. Even models considered "intrinsically interpretable" may have hidden vulnerabilities, for example a recent stody on PIP-Net and Proto-ViT models \cite{baniecki_birds_2025} demonstrated that a prototype-based classifier could be deceived by subtle input manipulations, causing a bird image to be “explained” using car prototypes and thus undermining the model’s reasoning transparency. Such examples contradict the notion that interpretability alone guarantees robustness. Likewise, large language models (LLMs) may exploit latent linguistic patterns to influence users: recent studies such as \cite{mieleszczenko-kowszewicz_dark_2024} exposed how LLMs adjust their style to a reader’s Big Five personality traits, effectively producing personalized persuasive (or manipulative) outputs that could go unnoticed without adversarial scrutiny. More broadly, recent studies emphasize that explanation tools themselves can be targets of attack. A survey \cite{baniecki_adversarial_2024} catalogs numerous ways to manipulate or “fairwash” model explanations – for instance, by tweaking inputs or training data to hide biased reasoning – and it reviews defense strategies to secure explainability methods.  These findings underscore the need to rigorously evaluate and challenge both models and their explanations. 

\textbf{Right for the Wrong Reason} AI models can sometimes achieve high accuracy with predictions that are right for the wrong reason, meaning they make correct predictions but rely on spurious or misleading patterns rather than genuine, generalizable features. This issue is epitomized by the \textit{Clever Hans} effect, named after a horse that appeared to perform arithmetic by picking up on its trainer’s cues. For example, the \textit{Unmasking Clever Hans} paper \cite{lapuschkin_unmasking_2019} showed that an image classifier could correctly recognize horses not by identifying horse-specific visual features, but by latching onto incidental cues like a copyright tag frequently present in horse photos. Similarly,  the LIME paper \cite{ribeiro_why_2016} used their new interpretability technique to reveal that a seemingly accurate “wolf vs. husky” classifier was basing its decisions on the presence of snow in the background of the images — effectively detecting the background context rather than the animals themselves. These are prime examples of a model being right (getting the correct class) for the wrong reason (using a shortcut that coincides with the correct answer in the training data). While the above instances involve supervised classifiers, recent work indicates that unsupervised learning models are also vulnerable to such Clever Hans-style behavior. Recent studies \cite{kauffmann_explainable_2025} demonstrate that even state-of-the-art unsupervised representation learning methods can form spurious associations, for instance grouping or explaining images by irrelevant features like embedded text logos or high-frequency noise patterns. In other words, an unsupervised model might “explain” or organize data in a way that accidentally aligns with the task at hand, yet fails when those coincidental cues change. 

These findings demonstrate a recurring pattern: foundation models, and AI models in general, perform worse than initially reported when subjected to targeted, context-specific validation. High-stakes domains require not only standard benchmark evaluation but also rigorous verification frameworks is therefore crucial for responsible adoption of foundation models, particularly in healthcare, legal systems, education, and other safety-critical sectors.

\textbf{Model Evaluation Levels}
AI models are being applied to more and more important decisions, thus ensuring the reliability and trustworthiness of AI models stands as a critical endeavor. Model validation---an essential step in this process---serves as the gatekeeper between well performing and faulty model.
Regardless of which metrics are used to evaluate a model, it is critical to define the level at which the evaluation is conducted.  In this section, we propose five levels of model evaluation, highlighting how rarely models are examined at the higher levels.

\begin{figure}
    \centering
     \includegraphics[width=0.45\textwidth]{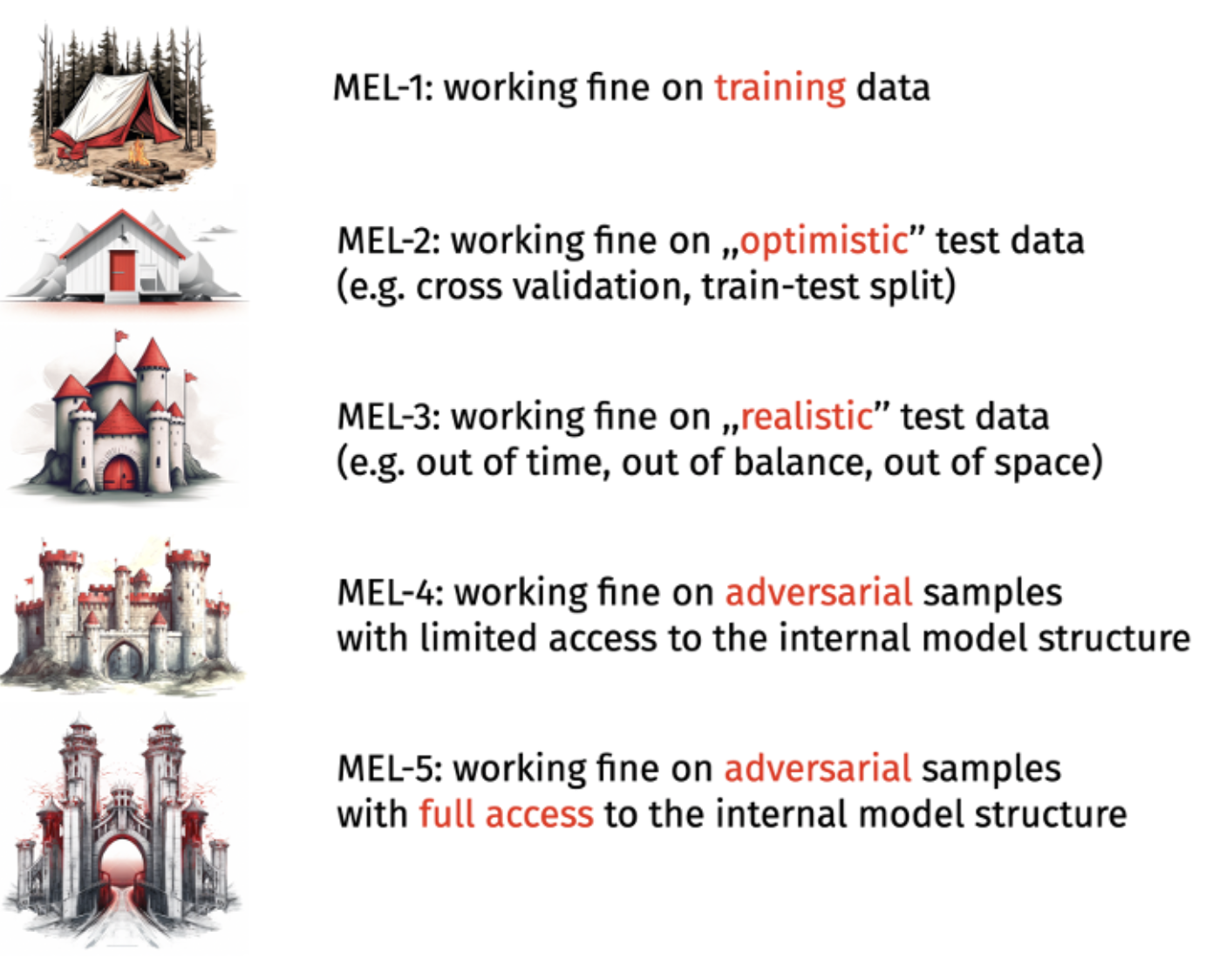}
     \caption{Graphical summary of the five levels of model evaluation}
    \label{fig:mel_model_science}
\end{figure}

When evaluating a model, both the evaluation criterion and the selection of the data sample to assess the quality of the model are important. In this section, we will look at the issue of data used for validation. Figure \ref{fig:mel_model_science} provides a graphical summary for model evaluation levels.

\textbf{Model Evaluation Level 0} --- no explicit model evaluation. It also includes the use of models for a new task for which they were not designed (unintended use) without any verification.

\textbf{Model Evaluation Level 1} --- for model evaluation we use the same data on which we trained the model. This is a common practice in statistical modelling based on linear models or other models where we have strong assumptions about the mechanism generating the data and the representativeness of the sample. For linear models, evaluation can be based on $R^2$ or p-values for the significance of the coefficients, typically calculated on the training sample.

\textbf{Model Evaluation Level 2} --- to evaluate the model, we use a separate data set with a distribution similar to the training data. The most common scenario is to randomly divide the data into two parts training and testing. Sometimes used in more sophisticated variants such as cross-validation or an out-of-bag data for bootstrap sampling. The characteristic of this evaluation is that the distribution of training and test data is close, which corresponds to the assumption that the reality in which the model will operate is not different from the reality in which the model was trained. This assumption definitely simplifies model validation, although covid pandemic, wars or other impactfull events, show that future data can differ from past data.

\textbf{Model Evaluation Level 3} --- to evaluate the model, ``disjoint'' data is used for training and testing. If the data contains a timestamp, the disjointness may mean using a different time range (so called out-of-time validation \cite{ifrs9_2021}). If the data contains a geographic tag, validation may mean using a different region (out-of-region validation). Virtually any attribute can be used to define a test sample, usually this is done to test for generalizability. For data collected from different devices, out-of-device validation can be considered, for AutoML systems working for many datasets, the out-of-dataset validation \cite{ho_extensions_2020} and so on.
This testing scheme is used by audit teams in (for example) financial institutions, especially for large-scale, high impact models, such as credit scoring models.

\textbf{Model Evaluation Level 4} --- it is time to actively look for cases in which the model does not work, even if they are rare according to the training data. There are situations in which users of the model may want to adversarially influence the model's decisions. For example, for an insurance pricing model, the user is tempted to manipulate certain parameters so as to evaluate the price of the premium. Parameters such as mileage, a statement about the number of drivers or number of days being abroad may be given incorrectly by the user if he or she thinks they will reduce the price of the premium. Another example is models that detect hate speech in online forums, users who want to post offensive comments will actively test scenarios in which the models do not work. In these situations, the adversary does not know the structure of the model, and can only reflect it and thus look for model vulnerabilities.

\textbf{Model Evaluation Level 5} --- In a more extreme scenario, the adversary has full access to the model. Whether it has been stolen or the model is publicly available is of little relevance. Analysis at this level allows tracking decision paths, calculating and using gradients. An example is the use of the Llama 2 \cite{llama2} language model, which is publicly available, in an AI system, or the Gandalf Challenge\footnote {\url{https://gandalf.lakera.ai/}} designed to deceive LLM models in order to steal a secret, i.e., perform undesirable behavior.

\textbf{Standards and Checklists} How to reach higher MEL? Clear verification procedures, standards, and checklists are essential for building trustworthy AI systems. Studies on medical imaging models for COVID-19 show that many errors came from poor data handling and weak validation, which led to misleading results; a structured checklist was proposed to guide responsible model development in such high-stakes domains \cite{hryniewska_checklist_2021}. In finance, transparency and auditability are critical for credit scoring models, and without clear validation rules, advanced algorithms cannot be trusted by regulators \cite{bucker_transparency_2022}. With the rapid growth of generative AI, experts call for new standards to protect the information ecosystem: public release of generative models should require built-in detection mechanisms \cite{knott_generative_2023}. Media and technology companies are also expected to follow new obligations to detect AI-generated content and ensure responsible communication \cite{knott_ai_2024}. These procedures, tools, and regulations are not just formal requirements; they are key to making AI systems safe, transparent, and reliable in sensitive areas such as healthcare, finance, and public communication.

\section{Explanation: how it really works?}

In recent years, the field of explainable machine learning has been booming, resulting in numerous papers proposing new methods and applications of these methods in relevant areas, both from the perspective of the end user (BLUE-XAI) and the model developer (RED-XAI) \cite{biecek24position}.
Below, we review some advances in interpretability and explainability, focusing on (i) representation and attention analysis, (ii) feature attribution, (iii) knowledge extraction, (iv) case-based explanations, and (v) model auditing (red teaming).

This overview does not replace a systematic surveys (such as \cite{Holzinger2022,guidotti_survey_2019,biecek_explanatory_2021}), but aims to show the synergies between the perspectives used in explanatory model analysis and in Model Science, which is more focused on exploration of specific models. Let's see how this is relevant for foundation models, such as {CLIP} for language-image pairs, {LLaMA} for language, {AlphaFold} for genetic sequences, or {AlphaZero} for chess-like games.

\textbf{Representation and Attention Analysis}. \textbf{CLIP use case.} Several studies decomposed CLIP's image-text representations to discover human-interpretable concepts. For example, \cite{gandelsman_interpreting_2024} revealed that individual attention heads specialize in distinct semantic roles (e.g., object localization, shape detection), enabling the removal of spurious features and even zero-shot segmentation. Sparse coding methods such as SpLiCE~\cite{bhalla_interpreting_2024} and Matryoshka Sparse Autoencoders (MSAE)~\cite{zaigrajew_interpreting_2025} expressed CLIP embeddings as a small set of semantic basis vectors, exposing $>100$ latent concepts useful for bias detection.
\textbf{Large Language Models (LLMs)  use case.} Mechanistic interpretability identified functional \emph{sub-circuits} in transformers. Gould et al.~\cite{gould_successor_2023} described \emph{successor heads} responsible for incrementing sequences (e.g., “Monday” $\rightarrow$ “Tuesday”) across multiple LLM families. Causal Head Gating (CHG)~\cite{nam_causal_2025} automatically classified attention heads as helpful, disruptive, or redundant, highlighting low modularity and overlapping circuits in LLaMA.
\textbf{AlphaFold  use case.} Analyses showed that Evoformer attention mechanisms capture \emph{coevolutionary couplings}, similar to Direct Coupling Analysis in bioinformatics~\cite{lupo_protein_2022}. Simplified single-layer attention models can recover contact maps, suggesting AlphaFold implicitly learns evolutionary constraints.
\textbf{AlphaZero  use case.} Probing studies revealed that its value network encodes classical chess concepts (e.g., king safety, pawn structure) despite training without supervision~\cite{mcgrath_acquisition_2022}. Furthermore, ~\cite{palsson_unveiling_2023} extracted novel strategic concepts from AlphaZero, successfully transferring them to human grandmasters.

\textbf{Feature Attribution}
Feature attribution moved from raw inputs to latent representations. For CLIP, \cite{dreyer_what_2025} combined sparse autoencoders and \emph{attribution patching} to quantify the influence of each latent component, revealing hidden biases (e.g., textual artefacts strongly influencing predictions). In LLMs, causal ablations, activation patching, and influence functions traced specific neurons and training examples affecting a given output~\cite{meng_locating_2023}. ExplainableFold~\cite{tan_explainablefold_2023} introduced \emph{counterfactual explanations} for AlphaFold by generating hypothetical sequence mutations altering the predicted 3D structure, resembling in-silico biological experiments.

\textbf{Case-Based Explanations}
Case-based XAI helps interpret single predictions. CLIP-InterpreT~\cite{gandelsman_interpreting_2024} provides interactive per-head nearest neighbors, showing which attention heads focus on which concepts. In AlphaFold, pLDDT confidence scores and sequence-level contributions (MSA evidence) guide biologists in assessing reliability. For AlphaZero, active neurons linked to strategic motifs (e.g., “passed pawn”) offer human-readable rationales for specific moves.

\textbf{Exploring with synthetic data} Synthetic samples, may be use to probe and explain black-box models. A popular scheme for such exploration are so colled counterfactual explanations, that adress the question “what if?” and generating plausible data variations that reveal the causal factors behind predictions. Recent work extends this paradigm to different modalities and scales. For time-series, \cite{pludowski_mascots_2025}  introduce \textit{MASCOTS}, a model-agnostic method that operates in a symbolic space to replace influential subsequences with minimal changes, ensuring both fidelity and interpretability of counterfactuals while highlighting pivotal temporal patterns in classification decisions. In the vision domain, \cite{sobieski_rethinking_2024} proposes region-constrained visual counterfactuals, restricting alterations to user-defined regions, which yields clearer causal evidence and prevents misleading global modifications. Moving beyond local explanations, \cite{sobieski_global_2025} introduces \textit{Global Counterfactual Directions}, latent-space transformations that consistently flip predictions across many instances, providing a global view of decision boundaries and exposing systemic biases. Together, these approaches demonstrate that generating synthetic observations is not merely a post-hoc explanation tool but a stress-testing mechanism that enhances transparency, robustness, and trust. 

\textbf{Auditing and Red Teaming}
Systematic \emph{red teaming} exposes failure modes. Anthropic’s large-scale red teaming~\cite{ganguli_red_2022} showed that RLHF-trained LLMs become increasingly robust to harmful prompts with scale, while “raw” models remain vulnerable. Vision-language models are susceptible to adversarial patches exploiting global attention in ViTs~\cite{lovisotto_give_2022}, whereas AlphaFold overconfidently predicts plausible but incorrect folds in proteins lacking evolutionary homologs~\cite{wang_overview_2024}. Identifying such weaknesses informs model debugging and safety improvements.

\begin{figure}
    \centering 
    \includegraphics[width=0.45\textwidth]{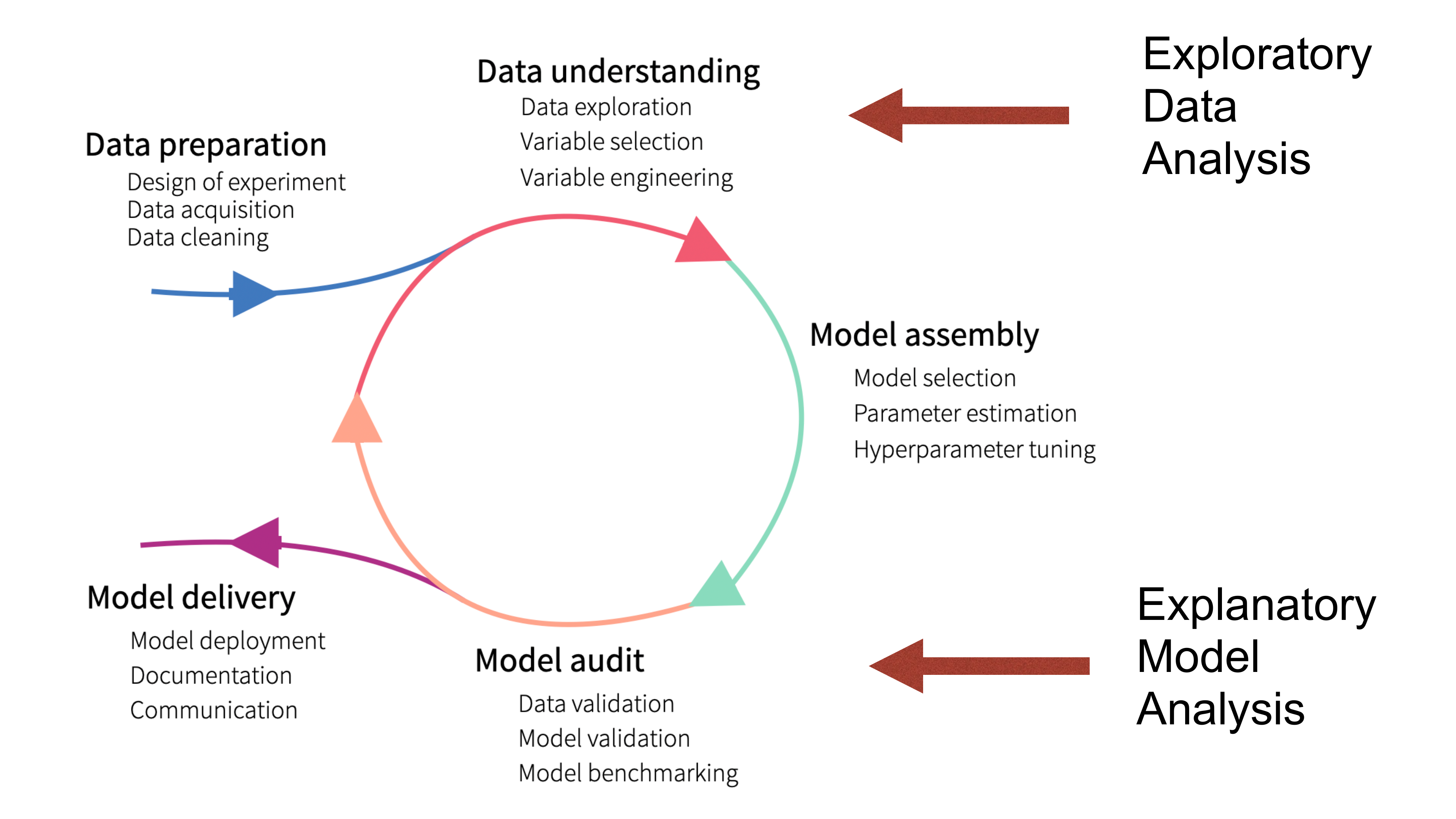}
    \caption{Explanatory model analysis can play a similar role to exploratory data analysis. The difference is that it takes place at a different part of the model training cycle.}
    \label{fig:waves_model_science2}
\end{figure}

Together, these advances demonstrate that complex foundation models can be probed, decomposed, and audited to extract meaningful scientific and strategic knowledge. From mechanistic circuits in LLaMA to novel chess strategies in AlphaZero, XAI research evolves from \emph{justifying} predictions to \emph{questioning and understanding} models as knowledge-bearing systems. Figure \ref{fig:waves_model_science2} shows that explainable model analysis can be an integral part of the model development process.

\section{Control: how it should work?}

Recent advances in alignment research have substantially improved the reliability of large language models (LLMs) and multimodal models (e.g., CLIP). Among the recent studies, {Ouyang et al. (2022)} introduced \textit{InstructGPT}~\cite{Ouyang2022}, which combined supervised fine-tuning (SFT) on high-quality instruction-response pairs with {Reinforcement Learning from Human Feedback (RLHF)}. This approach significantly reduced hallucinations and toxic outputs, establishing RLHF as a standard for aligning LLMs.

Building on this, \textit{Constitutional AI}~\cite{bai_constitutional_2022} replacing most human feedback with AI-generated critiques guided by a set of ethical principles. This {RL from AI Feedback (RLAIF)} achieved comparable safety while scaling alignment to larger models at lower cost.
To address the complexity of RLHF, \textit{Direct Preference Optimization (DPO)} was introduced~\cite{rafailov_direct_2024}, a closed-form supervised objective that aligns models directly on human preference data, matching RLHF performance without training separate reward models. Similarly, {Lee et al. (2024)} empirically validated RLAIF vs. RLHF~\cite{lee_rlaif_2024}, confirming that AI feedback can substitute human preference data for many tasks.

Recent methods directly extracted symbolic or conceptual knowledge. In LLMs, ROME and MEMIT~\cite{meng_locating_2023,meng_mass_editing_2023} demonstrated that factual triples (\textit{subject--relation--object}) are stored in localized MLP weights and can be edited or retrieved. CLIP’s semantic basis vectors serve as a \emph{conceptual vocabulary}, enabling bias audits~\cite{zaigrajew_interpreting_2025}. For reinforcement learning, NUDGE~\cite{delfosse_interpretable_2023} distilled logical rules from neural policies, a potential path to symbolic AlphaZero strategies.

Finally, multimodal alignment gained attention. {Zhou et al. (2023)} demonstrated universal adversarial attacks on CLIP (\textit{AdvCLIP})~\cite{zhou_advclip_2023}, prompting robust fine-tuning strategies for vision encoders. Recent extensions (e.g., layer-wise PPO on VLMs) adapt RLHF-style techniques to vision-language models, reducing harmful outputs triggered by adversarial images.

AI model control is still a developing field, responding both to the different expectations of different user groups (the need to personalize model responses) and to the growing number of identified model misbehaviours. This is a natural continuation of the verification and explanation pillars, because once we know that something is not working and understand why, the next step is to apply mechanisms to improve model behaviour, preferably without complete retraining.

\section{Interface: how to interact with it?}

Foundation models such as large language models and vision transformers have achieved unprecedented performance on diverse tasks, but their internal reasoning remains largely opaque to users \cite{bommasani_opportunities_2022}. This opacity raises concerns about trust and accountability, especially as these models are deployed in high-stakes domains. To address this, researchers are exploring explanation interfaces that leverage interaction and visualization to help humans comprehend complex model behavior. Rather than treating a foundation model as a “black box,” interactive interfaces aim to open up the model’s decision process, allowing users to probe \emph{why} a model produced a given output and under what conditions it might change its decision \cite{miller_explanation_2019}. These approaches build on the notion that effective explanations can improve users’ mental models of the AI system and calibrate their trust accordingly \cite{kulesza_principles_2015}.

\textbf{Model exploration}
An important challenge in model exploration is combining multiple explanatory methods into coherent, user-driven workflows. The \textit{Grammar of Interactive Explanatory Model Analysis} (IEMA) formalizes such workflows, enabling sequential exploration of diverse methods and supporting structured human--model dialogues \cite{baniecki_grammar_2023}. Such multi-perspective analysis improves understanding and decision-making but requires careful design to avoid overwhelming users.
Another challenge concerns aligning explanations with human semantics and enabling user guidance. Methods like \textit{LIMEcraft} allow analysts to craft meaningful superpixels and inspect how predictions change, revealing model biases that purely algorithmic segmentation might miss \cite{hryniewska_limecraft_2022}. Similarly, animated visualizations offer richer insight into complex behaviors: Spyrison et al.\ demonstrate that dynamic linear projections can expose local decision boundaries of nonlinear models, supporting intuitive “what-if” analyses \cite{spyrison_exploring_2025}.
Finally, conversational interaction is emerging as a promising paradigm. Studies on human–model dialogues show that users ask diverse, context-dependent questions—about reasons, counterfactuals, and uncertainties—that static tools often fail to address \cite{kuzba_what_2020}. Building systems that respond naturally and faithfully to such queries remains an open problem.

\textbf{Interactive explanation tools} enable a two-way exchange between the user and the model, transforming explanation from a static output into an ongoing dialogue. One line of work integrates explanations into chat-based interactions with large language models. For instance, an LLM can justify its answer or clarify specific steps when asked, providing on-demand self-explanations. More structured interfaces allow users to \emph{manipulate} model inputs or reasoning and see the effects. Kulesza et al.’s early concept of \textit{explanatory debugging} highlighted how users could iteratively correct a model by giving feedback on its explanations \cite{choo_visual_2018}. Modern interfaces are bringing this vision to foundation models. For example, \cite{wu_promptchainer_2022} introduced \textit{PromptChainer}, a visual interface for building and debugging chains of LLM prompts. It lets users break a complex task into a sequence of prompt transformations, making the overall reasoning process more transparent and controllable. Other systems support interactive prompt refinement: \cite{strobelt_interactive_2022} lets users visually edit and experiment with prompt phrases and observe changes in the LLM’s output in real time. Similarly, \textit{ScatterShot} is an interface that helps users curate in-context examples for few-shot learning by interactively selecting and revising examples, which in turn updates the model’s behavior \cite{wu_scattershot_2023}. By allowing users to test “what if” scenarios—e.g. “What would the model do if we tweak this prompt or example?”—these interfaces help users identify the model’s decision boundaries and failure modes. Such iterative exploration can not only improve task performance but also give users a sense of agency and deeper understanding of the model.

\textbf{Visualization-based explainability} plays a complementary role, especially for vision and transformer-based models. Complex models often produce high-dimensional internal states that defy simple explanation; visualization techniques translate these into human-interpretable forms. For vision foundation models, spatial heatmaps and feature visualizations are common: e.g. Grad-CAM highlights which regions of an image influence a CNN’s prediction \cite{selvaraju_grad-cam_2017}, helping users verify if the model is “looking” at relevant features (like the suspected tumor area in a medical image). Such visual explanations can be integrated into interfaces where a user hovers over or clicks on parts of an input to reveal the model’s attention or confidence for those parts. \cite{derose_attention_2020} present attention flow diagrams that trace how information propagates between words across the layers of a transformer. Their interface allows users to select a token and see which other tokens it attends to and influences, illuminating the model’s latent reasoning chain. Other tools like \textit{AttentionViz} \cite{yeh_attentionviz_2023} provide a global view of attention patterns, summarizing where the model focuses across multiple inputs and layers. This helps in spotting broad trends, such as whether a language model disproportionately attends to certain prompt keywords or if a vision transformer’s attention aligns with meaningful image regions \cite{li_how_2023}. By interacting with these visualizations (for example, filtering attention by layer or threshold, or comparing attention before and after fine-tuning), users can investigate model behaviors at different levels of granularity. 

\textbf{User trust and understanding} are central metrics for the success of these model exploration interfaces. The goal is not only to inform the user, but to do so in a way that calibrates their trust—users should trust the model when it is correct and be appropriately cautious when the model is uncertain or makes an error. Interactive and visual explanations can aid in this calibration by revealing the rationale behind model outputs. Recent studies shows that poor interface design or misleading explanations can backfire. \cite{kaur_interpreting_2020} found that data scientists often misinterpreted outputs from interpretability tools and became overconfident in the model’s correctness even when the explanations were superficial. This highlights that simply providing an explanation interface is not enough; the explanations themselves must be faithful and understandable. \cite{bansal_does_2021} similarly showed that while explanations (such as highlighting important features or providing model confidence) can improve human-AI team decision performance in some cases, they can also decrease performance if the model is wrong and the explanation leads the human to over-rely on the AI. To build trustable interfaces, designers are adopting UX approaches: conducting user studies to see how people actually use and react to explanations, and iterating on the interface to better support users’ decision-making and sensemaking processes. In some cases, this means allowing users to challenge the model’s reasoning (e.g. by asking “why not?” questions or providing counterexamples) and having the system respond with revised explanations or outputs, in others, it means clearly conveying model uncertainty.

\section{Conclusions}

We are witnessing the emergence of new AI models that are continuously trained on diverse datasets, which are often no longer accessible afterwards. These models will become the foundation for numerous tools that rely on their representations in relation to dozens of other important tasks. It is therefore crucial to develop methods, processes, and exemplary analyses of such models, addressing both their efficiency and safety.

In this article, we introduced the concept of Model Science—a subfield of research focused on the analysis of models. We identified four key areas essential for such analysis and outlined initial research directions already underway in these domains.
By providing a structured framework for this field, we hope to foster the further development of methods for analyzing models.
We also invite readers to engage in both shaping and promoting the field of Model Science.

\begin{ack}
Work on this project is financially supported from the SONATA BIS grant 2019/34/E/ST6/00052 funded by Polish National Science Centre (NCN) and INFOSTRATEG-I/0022/2021-00 grant funded by Polish National Centre for Research and Development (NCBiR).
\end{ack}

%%%%%%%%%%%%%%%%%%%%%%%%%%%%%%%%%%%%%%%%%%%%%%%%%%%%%%%%%%%%%%%%%%%%%%%%

%%% Use this command to include your bibliography file.

\bibliographystyle{apalike}
\bibliography{mybibfile}

\end{document}